# Machine Learning of User Profiles: Representational Issues

[+][*]Eric Bloedorn, [+]Inderjeet Mani, and [+]T. Richard MacMillan

[+]Artificial Intelligence Technical Center, The MITRE Corporation
Z40,11820 Dolley Madison Boulevard, McLean, VA 22102
{bloedorn, imani, macmilla}@mitre.org

[*]Machine Learning and Inference Laboratory
George Mason University, Fairfax, VA 22030

**Abstract**

As more information becomes available electronically, tools for finding information of interest to users becomes increasingly important. The goal of the research described here is to build a system for generating comprehensible user profiles that accurately capture user interest with minimum user interaction. The research described here focuses on the importance of a suitable generalization hierarchy and representation for learning profiles which are predictively accurate and comprehensible. In our experiments we evaluated both traditional features based on weighted term vectors as well as subject features corresponding to categories which could be drawn from a thesaurus. Our experiments, conducted in the context of a content-based profiling system for on-line newspapers on the World Wide Web (the IDD News Browser), demonstrate the importance of a generalization hierarchy and the promise of combining natural language processing techniques with machine learning (ML) to address an information retrieval (IR) problem.

## Introduction

As more information becomes available on the Internet, the need for effective personalized information filters becomes critical. In particular, there is a need for tools to capture profiles of users' information needs, and to find articles relevant to these needs, as these needs change over time. Information filtering, as (Belkin & Croft 1992), (Foltz & Dumais 1992) point out, is an information access activity similar to information retrieval, but where the profiles represent evolving interests of users over a long-term period, and where the filters are applied to dynamic streams of incoming data.

Our research builds on two particular traditions involving the application of machine learning to information access: empirical research on relevance feedback within the information retrieval community, and interdisciplinary work involving the construction of personalized news filtering agents. We will now introduce these briefly, to better motivate and distinguish our work.

Relevance feedback approaches are a form of supervised learning where a user indicates which retrieved documents are relevant or irrelevant. These approaches, e.g., (Rocchio 1971), (Robertson & Sparck-Jones 1976), (Belew 1989), (Salton & Buckley 1990), (Harman 1992), (Haines & Croft 1993), (Buckley, Salton, & Allan 1994), have investigated techniques for automatic query reformulation based on user feedback, such as term reweighting and query expansion. While this body of work is not necessarily focused exclusively on the information filtering problem, it demonstrates effectively how learning can be used to improve queries.

Work on the application of machine learning techniques for constructing personalized information filters has gained momentum in recent years. Filters have been constructed for many applications including web pages and USENET news, such as NewT (Maes 1994), Webhound (Lashkari, Metral, & Maes 1994), WebWatcher (Armstrong et al. 1995), WebLearner (Pazzani et al. 1995) and NewsWeeder (Lang 1995).

However, in the above approaches, the agent is unable to learn *generalizations* about the user's interests. For example, if a user likes articles on *scuba*, *whitewater rafting*, and *kayaking*, an agent with the ability to generalize could infer that the user is interested in *water sports*, and could communicate this inference to the user. Not only would this be a natural suggestion to the user, but it might also be useful in quickly capturing their real interest and suggesting what additional information might be of interest. Thus, generalization could provide a powerful mechanism for capturing and communicating representations of user's interests, as well as justifying recommendations. More importantly the description provided by the generalization could help make the profile more intelligible to the user, which in turn could help establish trust.

Such an approach could exploit a concept hierarchy or network to perform the generalizations. While thesauri and other conceptual representations have been the subject of extensive investigation in both query formulation and expansion (e.g., see (Jones et al. 1995) for detailed references), they have not been used to learn generalized profiles. In this paper, we describe a profile representation which exploits (together with other features) summary-level features, extracted from text using language processing techniques, linked to a generalization hierarchy from a thesaurus.

We report on experiments evaluating the effect of various document representations on profile learning with a variety of learning algorithms. Our experiments were conducted in the context of a content-based profiling and summarization system for on-line newspapers on the World Wide Web, the IDD News Browser. In this system the user can set up and edit profiles, which are periodically run against various collections built from live Internet newspaper and USENET feeds, to generate matches in the form of personalized newspapers. These personalized newspapers provide multiple views of the information space in terms of summary-level features. When reading their personalized newspapers, users provide positive or negative feedback to the system, which are then used by a learner to induce new profiles. These system-generated profiles can be used to make recommendations to the user about new articles and collections.

## Text Representation

The basic goal of building a summary-level representation of text is to build a rich characterization of the document text that highlights those aspects which are important over those which are not. This summary-level encoding allows for a more efficient representation, potentially reducing the dimensionality of the feature spaces used in learning. Dimensionality reduction (e.g., Latent Semantic Indexing (Deerwester et al., 1990)) has been a long-standing issue in information retrieval, where it is not uncommon to consider $10^5$ or more features due to the large vocabulary of a text collection.

Our first set of summary level features assigns thesaural categories (or subjects) to segments of text based on the Subject Field Coder (SFC) (Liddy and Paik 1992) (Liddy and Myaeng 1992) (from TextWise, Inc.). In attempting to associate thesaural categories with texts, one well-known problem is that of word-sense disambiguation, in this case deciding which of several thesaurus categories are the most likely ones for a term in the text. The SFC approach exploits evidence from local context (e.g., unambiguous words) along with evidence from large-scale statistics (e.g., pairwise correlations of subject categories) to disambiguate word sense, producing a vector representation of a text's subject categories.[1] Text summaries are represented by vectors in 124-dimensional space, with each vector's projection along a given dimension corresponding to the salience in the text of that subject category.

The generalization hierarchy, which came to us from TextWise Inc.'s thesaurus consists of three levels. The SFC subject attributes provide values for individual documents at the lowest level (level 3) of the hierarchy. Although this hierarchy covers a fairly wide set of subjects as required for our newspaper application, it does not have a great deal of depth to allow more specific profiles to be differentiated. To remedy this, we extended the set of terminal categories under one of the categories, medicine, by hand to include another 16 lowest level categories.

Based on recent findings by (Broglio & Croft 1993), on improved retrieval performance on TIPSTER queries, and to further strengthen profile intelligibility, we included features in our document representation involving terms relating to People, Organizations, and Locations (POLs) (along with their respective attributes). These features were provided by a name tagger (Mani & MacMillan 1995) which classifies names in unrestricted news articles in terms of a hierarchy of different types of POLs along with their attributes (e.g., a person's title, an organization's business, a city's country, etc.) The name tagger combines evidence from multiple knowledge sources, each of which uses patterns based on lexical items, parts of speech, etc., to contribute evidence towards a particular classification.[2]

Document keywords were also extracted by using a term-frequency inverse-document-frequency (tf.idf) calculation (Salton & McGill 1983), which is a well-established technique in information retrieval. The weight of term *k* in document *i* is represented as:

$$dw_{ik} = tf_{ik} * (\log_2(n) - \log_2(df_k) + 1)$$

Where $tf_{ik}$ = frequency of term k in document i, $df_k$ = number of documents in which term k occurs, n = total number of documents in collection. Given these three sources of features, we developed a hybrid document representation. Features describe subjects (x1..x5), people (x6..x59), organizations (x60..x104) locations (x105..x140) and the top n statistical keywords (x140..x140+n), where n was varied from 5 to 200. Attributes (x1..x5) are linked to the thesaural generalization hierarchy.

## Learning Methods

Our representational decisions suggested some constraints on the learning method. We wanted to use learning methods which performed inductive generalization where the SFC generalization hierarchy could be exploited. Also, we required a learning algorithm whose learnt rules could be made easily intelligible to users. We decided to try both AQ15c (Wnek et al. 1994) and C4.5-Rules (Quinlan 1992) because they meet these requirements (the generalization hierarchy is made available to C4.5 by extending the attribute set), are well-known in the field and are readily available.

---

[1] An earlier version of the SFC was tested on 166 sentences from the Wall Street Journal (1638 words) giving the right category on 87% of the words (Liddy and Myaeng 1992).

[2] In trials against hand-tagged documents drawn from collections of newswire texts, the tagger had an average precision-recall of approximately 85%.

| Metric | Definition |
| --- | --- |
| Predictive Accuracy: | # testing examples classified correctly / total number of test examples. |
| Precision: | # positive examples classified correctly / # examples classified positive, during testing |
| Recall: | # positive examples classified correctly / # known positive, during testing |

Figure 1. Metrics used to measure learning performance

AQ15c is based on the $A^q$ algorithm for generating disjunctive normal form (DNF) expressions with internal disjunction from examples. In the $A^q$ algorithm rule covers are generated by iteratively generating stars from randomly selected seeds. A star is a set of most general alternative rules that cover that example, but do not cover any negative examples. A single 'best' rule is selected from this star based on the user's preference criterion (e.g. maximal coverage of new examples, minimal number of references, minimal cost, etc.). The positive examples covered by the best rule are removed from consideration and the process is repeated until all examples are covered. AQ15c makes use of a generalization hierarchy in an attribute domain definition by climbing the hierarchy during rule construction.

C4.5-Rules, which is part of the C4.5 system of programs, generates rules based on decision trees learned by C4.5. In C4.5 a decision tree is built by repeatedly splitting the set of given examples into smaller sets based on the values of the selected attribute. An attribute is selected based on its ability to maximize an expected information gain ratio. In our experiments we found the pruned decision rules produced the most accurate predictions.

An advantage of experimenting with these two learning methods is that we get to see if the representation we have developed for this problem is providing a useful characterization of the text, or if it is just well-matched to the bias of the selected learning algorithm. If the former, our results are possibly applicable to other applications of machine learning to text.

## Experimental Design

The goal of these experiments was to evaluate the influence of different sets of features on profile learning. In particular, we wanted to test the hypothesis that summary level features used for generalization were useful in profile learning. Each of the experiments involved selecting a source of documents, vectorizing them, selecting a profile, partitioning the source documents into documents of interest to the user (positive examples) and not of interest (negative examples), and then running a training and testing procedure. The training involved induction of a new profile based on feedback from the pre-classified training examples. The induced profile was then tested against each of the test examples. One procedure used 10 runs in each of which the examples were split into 70% training and 30% test (70/30-split). Another procedure used a 10-fold cross-validation, where the test examples in each of the 10 runs were disjoint (10-fold-cross). The metrics we used to measure learning on the USMED and T122 problems include both predictive accuracy and precision and recall. These metrics are defined as shown in Figure 1. Precision and recall are standard metrics in the IR community, and predictive accuracy is standard in the ML community.

Our first experiment exploited the availability of users of the IDD News Browser. A user with a "real" information need was asked to set up an initial profile. The articles matching his profile were then presented in his personalized newspaper. The user then offered positive and negative feedback on these articles. The set of positive and negative examples were then reviewed independently by the authors to check if they agreed in terms of relevance judgments, but no corrections needed to be made. The details of the test are[3]:

Source: Colorado Springs Gazette Telegraph (Oct. through Nov. 1994) Profile: "Medicine in the US" (USMED) Relevance Assessment: users, machine aided Size of collection: 442 Positive Examples: 18 Negative Examples: 20 Validation: "70/30-split"

Our next experiment exploited the availability of a standard test collection, the TREC-92 collection. The same generalization hierarchy used in the previous experiment was used here. The idea was to study the effect that these changes in the hierarchy would effect learning of the other topics. The details of the test are:

Source: Wall Street Journal (1987-92), Profile: "RDT&E of New Cancer Fighting Drugs" (T122) Relevance Assessment: provided by TREC, Size of collection: 203, Positive Examples: 73, Negative Examples: 130, Validation: "10-fold cross"

## Experimental Results

In our first set of experiments we applied AQ15c and C4.5-Rules to the USMED and T122 datasets. Here AQ15c has the hierarchy available to it in the form of hierarchical domain definitions for attributes x1 through x5. C4.5 has a hierarchy available to it through an extended attribute set. In this extension, based on a pointer from Quinlan (Quinlan, 1995), we extended the attribute set to include attributes which describe nodes higher up on the generalization hierarchy. A total of eighteen additional attributes were added (six for each non-null subject attribute) which provided the values of the subject attributes at each of the six levels higher in the tree from the leaf node. Because the tree was unbalanced some of the additional attributes took dummy values for some examples.

---

[3] Feedback on only a small number of examples is quite typical of real-world applications.

| Learning Method | Learning Problem | Predictive Accuracy | | | | Average Precision/ Average Recall | | | |
|---|---|---|---|---|---|---|---|---|---|
| | | TFIDF | POL | SFC | ALL | TFIDF | POL | SFC | ALL |
| AQ15c | USMED | 0.58 | *0.48* | **0.78** | 0.55 | 0.51/1.00 | 0.45/0.45 | **0.78/0.73** | *0.52/0.34* |
| | T122 | *0.39* | 0.59 | 0.59 | **0.76** | 0.36/0.88 | 0.43/0.66 | *0.50/0.33* | **0.79/ 0.48** |
| C4.5-Rules | USMED | *0.39* | 0.74 | 0.79 | **0.76** | *0.07/0.30* | 0.89/0.60 | 0.97/0.60 | **0.90/0.60** |
| | T122 | *0.64* | 0.65 | 0.68 | **0.76** | *0.0/0.0* | 0.64./0.22 | 0.58 /0.55 | **0.70/ 0.67** |

Table 1. Predictive Accuracy, Average Precision, and Average Recall of learned profiles for a given feature set (averaged over 10 runs).

### Predictive Accuracy/ Precision and Recall

The predictive accuracy/precision and recall results (Table 1) show that the most predictively accurate profiles generated (boldface) come from either the SFC or ALL feature sets. The TFIDF scores are shown for *n*=5 (5 keywords); there was no appreciable difference for *n*=200. All differences between the best and worst predictive accuracies (in italics) are significant to the 90% level and were calculated using a student t-test.

These results suggest that profile learning using summary-level features (POL or SFC) alone or in combination with term-level features (TFIDF) provides useful information for characterizing documents. When a relevant generalization hierarchy is available, as witnessed by the superior performance of the SFC in the USMED, these summary features alone can be more predictively accurate than using only term-level features (TFIDF). When such background knowledge is not available, the hybrid representation was best for both C4.5-Rules and AQ15c. Our general conclusion is that these results reveal that the hybrid representation can be useful in profile learning.

### Intelligibility of Learnt Profiles

A system which discovers generalizations about a user's interests can use these generalizations to suggest new articles. However, as mentioned earlier, in many practical situations, a human may need to validate or edit the system's learnt profiles. Intelligibility to humans then becomes an important issue. The following profile induced by AQ illustrates the intelligibility property. It shows a generalization from terminal vector categories contagious and genetic present in the training examples to medical science, and from the terminal category abortion up to medical policy.

IF *subject1 = nature or physical science &
   subject2 = nature or medical science or medical policy
               or human body*
THEN article is of interest

C4.5 also produced many simple and accurate rules. One of the profiles generated by C4.5-Rules for the USMED dataset and ALL featureset shown below shows how C4.5 focused on the POL tags to come up with a simple description of the USMED articles in the training set.

IF *POLtag5_location = US  or  POLtag1_honorific = Dr.*
THEN article is of interest

Although intelligibility is hard quantify, we examined profile length, measured as the number of terms on the left hand side of a learnt rule. Here we observed that using ALL the features led to more complex profiles over time, whereas using only subsets of features other than POL leveled off pretty quickly at profiles with well under 10 terms. The SFC profiles, which exploited generalization, were typically short and succinct. The TFIDF profiles were also quite short, but given their low overall performance they would not be useful.

### Comparison with Word-Level Relevance Feedback Learning

In order to compare our results with a traditional relevance feedback method we applied a modified Rocchio algorithm to the two information retrieval tasks (USMED and T122) described earlier.

The modified Rocchio algorithm is a standard relevance feedback learning algorithm which searches for the best set of weights to associate with individual terms (e.g., tf-idf features or keywords) in a retrieval query. In these experiments individual articles are represented as vectors of 30,000 tf-idf features. Our Rocchio method is based on the procedure described in (Buckley, Salton, & Allan 1994). As before, the training involved induction of a new profile based on feedback from the pre-classified training examples, as follows. To mimic the effect of a user's initial selection of relevant documents matching her query, an initial profile was set to the average of all the vectors for the (ground-truth) relevant training documents for a topic. This average was converted from a tf.idf measure to a tf measure by dividing each tf.idf value by the idf. The profile was then reweighted using the modified Rocchio formula below. This formula transforms the weight of a profile term k from p-old to p-new as follows (Buckley, Salton, & Allan 1994):

$$p\text{-new}_k = (\alpha * p\text{-old}_k) + \left(\frac{\beta}{r} * \sum_{i=1}^{r} dw_{ik}\right) - \left(\frac{\gamma}{s} * \sum_{i=1}^{s} dw_{ik}\right)$$

| Learning Method | Predictive Accuracy | | Average Precision/ Average Recall | |
|---|---|---|---|---|
| | USMED | T122 | USMED | T122 |
| **Rocchio** | 0.49 | 0.51 | 0.52/0.53 | 0.39/0.27 |
| **Best AQ15c (SFC)** | 0.78 | 0.76 | 0.78/0.73 | 0.79//0.48 |
| **Best C4.5 (ALL)** | 0.76 | 0.76 | 0.97/0.60 | 0.70/0.67 |

Table 2. Comparing Predictive Accuracy, Average Precision / Average Recall for tf.idf terms.

Where $r$ = number of relevant documents, $dw_{ik}$ = tf weight of term $k$ in document $I$, $s$ = number of non-relevant documents, and the tuning parameters $\alpha = 8$, $\beta = 16$, and $\gamma = 4$. During testing, the test documents were compared against the new profile using the following cosine similarity metric for calculating the degree of match between a profile j (with the tf weights converted back to tf.idf weights) and a test document i (with tf.idf weights) (Salton & McGill 1983):

$$c_{ij} = \frac{\sum_{k=1}^{t} (dw_{ik} * qw_{jk})}{\sqrt{\sum_{k=1}^{t} dw_{ik}^2 * \sum_{k=1}^{t} qw_{jk}^2}}$$

Where $t$ = total number of terms in collection, $dw_{ik}$ = tf.idf weight of term $k$ in document $I$, $qw_{jk}$ = tf.idf weight of term $k$ in profile $j$. The cutoff for relevance was varied between 0 and 1, generating data points for a recall-precision curve. A best cutoff (which maximizes the sum of precision and recall) was chosen for each run. The results in Table 2 show that the machine learning methods represented by the best runs from AQ15c and C4.5 outperform the tf-idf based Rocchio method on both the T122 and USMED problems in terms of both predictive accuracy and predictive precision and recall.

## Conclusion

These results demonstrate that a relevant generalization hierarchy together with a hybrid feature representation is effective for accurate profile learning. Where the hierarchy was available and relevant, the SFC features tended to outperform the others, in terms of predictive accuracy, precision and recall, and stability of learning. Other features and combinations thereof showed different learning performance for different topics, further emphasizing the usefulness of the hybrid representation. These results also confirm the suspicion that tuning a thesaurus to a particular domain will generally yield better learning performance. In this connection, the work of (Evans et al. 1991) on thesaurus discovery and (Hearst & Schutze 1993) on thesaurus tuning is highly relevant. In the latter work, thesaural categories extracted automatically from Wordnet (Miller et al 1990) were extended with terms from a corpus. We can imagine a similar technique being used to augment the thesaurus used by the SFC.

Having assessed the basic performance of the profile learning capability, our next step will be to track the performance of the learner over time, where users of the IDD News Browser will have the option of correcting the induced profiles used to recommend new articles. We also hope to investigate the use of constructive induction to automate the search for an improved representation (Bloedorn & Michalski 1993).